\documentclass[conference]{IEEEtran}
\IEEEoverridecommandlockouts

\usepackage{cite}
\usepackage[T1]{fontenc}
\usepackage{amsmath,amssymb,amsfonts}
\usepackage{algorithmic}
\usepackage{graphicx}
\usepackage{textcomp}
\usepackage{xcolor}
\def\BibTeX{{\rm B\kern-.05em{\sc i\kern-.025em b}\kern-.08em
T\kern-.1667em\lower.7ex\hbox{E}\kern-.125emX}}

\begin{document}
\title{Corporate IT-support Help-Desk Process Hybrid-Automation Solution with Machine Learning Approach}
\author{\IEEEauthorblockN{Kuruparan Shanmugalingam}
\IEEEauthorblockA{\textit{Technology Alliances and Innovation} \\
\textit{Millennium I.T E.S.P (Pvt) Ltd}\\
Colombo, SriLanka \\
kuruparans@millenniumitesp.com}
\and

\IEEEauthorblockN{Nisal Chandrasekara}
\IEEEauthorblockA{\textit{Technology Alliances and Innovation} \\
\textit{Millennium I.T E.S.P (Pvt) Ltd}\\
Colombo, SriLanka \\
nisalc@millenniumitesp.com}
\and

\IEEEauthorblockN{Calvin Hindle}
\IEEEauthorblockA{\textit{Technology Alliances and Innovation} \\
\textit{Millennium I.T E.S.P (Pvt) Ltd}\\
Colombo, SriLanka \\
calvinh@millenniumitesp.com}
\and

\hspace{2.5cm}
\IEEEauthorblockN{Gihan Fernando}
\IEEEauthorblockA{\hspace{2.5cm}\textit{Technology Alliances and Innovation} \\
\hspace{2.5cm}
\textit{Millennium I.T E.S.P (Pvt) Ltd}\\
\hspace{2.5cm}Colombo, SriLanka \\
\hspace{2.5cm}
gihanf@millenniumitesp.com}
\and

\IEEEauthorblockN{Chanaka Gunawardhana}
\IEEEauthorblockA{\textit{Technology Alliances and Innovation} \\
\textit{Millennium I.T E.S.P (Pvt) Ltd}\\
Colombo, SriLanka \\
chanakag@millenniumitesp.com}

\thanks{First author is Kuruparan. All the authors are with Millennium I.T E.S.P (Pvt) Ltd. This is funded by Millennium I.T E.S.P (Pvt) Ltd.}}
\maketitle

\begin{abstract}
Comprehensive IT support teams in large scale organizations require more man power for handling engagement and requests of employees from different channels on a 24$\times$7 basis. Automated email technical queries help desk is proposed to have instant real-time quick solutions and email categorisation. Email topic modelling with various machine learning, deep-learning approaches are compared with different features for a scalable, generalised solution along with sure-shot static rules.  Email's title, body, attachment, OCR text, and some feature engineered custom features are given as input elements. XGBoost cascaded hierarchical models, Bi-LSTM model with word embeddings perform well showing 77.3 overall accuracy For the real world corporate email data set. By introducing the thresholding techniques, the overall automation system architecture provides 85.6 percentage of accuracy for real world corporate emails. Combination of quick fixes, static rules, ML categorization as a low cost inference solution reduces 81 percentage of the human effort in the process of automation and real time implementation.
\end{abstract}

\begin{IEEEkeywords}
Corporate emails, Feature engineering, Machine learning, Natural Language Processing, Robotic Process Automation,  Text classification, and Quick fixes
\end{IEEEkeywords}

\vspace{-0.5cm}
\section{Introduction}
In an organization, the Information Technology (IT) support help desk operation is an important unit which handles the IT services of a business. Many large scale organizations would have a comprehensive IT support team to handle engagement and requests with employees on a  24$\times$7 basis. As any routinized tasks, most  processes of the support help desk unit are considered repetitive in nature~\cite{park2019opportunities}. Some may occur on a daily basis and others may occur more frequently. Many support engineers and agent would spend time on these repetitive task such as entering information to an application, resetting passwords, unlocking applications, creating credentials, activating services, preparing documentation, etc.

The industry has now come realize that many repetitive business processes and tasks can be automated by using  Robotic Process Automation (RPA) bots or robotic processes automotive software bots \cite{al2019machine}. The idea is to take the repetitive workload and hand it over to the RPA bots so that the employees could focus on more value adding tasks and decision making to the organization. The RPA bot would also help to reduce the human errors and make processes more efficient, which would finally intent results in cost saving and productivity increase.

Our proposed automated approach is not only focused on automating repetitive tasks but also looking at historical data, enabling IT support desk process to identify unforeseen insights and patterns. Analyzing the data from various sources such as email communications, service request information generated from support ticketing applications and even conversational data from chats has helped us to identify the type of Service Requests (SR) raised and their respective solutions, as well as fixes done by the support agents. This approach has helped us create a classification model to identify the issue types and provide quick fixes and resolutions from the collected data.

\section{Related Work}
\emph{WrÃblewska} has conducted a project on the topic of RPA of unstructured data which was focused on  building an Artificial Intelligence (AI) system dedicated to tasks regarding the processing of formal documents used in different kinds of business procedures \cite{wroblewska2018robotic}. His approach was introduced to automate the debt collecting process. Possible applications of Machine Learning (ML) methods to improve the efficacy of these processes were described. In the case study done by \emph{Aguirre}, it was concluded that companies should consider RPA to be more suitable for high volume standardized tasks that are rule-driven, with no requirement for subjective judgement, creativity or interpretation skills \cite{aguirre2017automation}. Back office business processes such as accounts payable, accounts receivable, billing, travel and expenses, fixed assets and human resource administration are good candidates for RPA.

Extreme multi-class and multi-label text classification problems are solved by the methodology named Hierarchical Label Set Expansion (HLSE) \cite{gargiulo2019deep}. This paper presents the deep Learning architecture devoted to text classification, in which the data labels are regularized, the hierarchical label set is defined  and different word embeddings are used \cite{aguirre2017automation,stein2019analysis,kowsari2019text}.

The traditional model performed better than the the deep learning models for 8,841 emails collected over 3 years, because this particular classification task carried out by \emph{Haoran} may not require the ordered sequence representation of tokens that deep learning models provide \cite{zhang2019categorizing}. This paper claims that a bagged voting model surpasses the performance of any individual models.

In their survey, \emph{Kamran} and other researchers analyzed text feature extractions \cite{masood2019cognitive,huang2019efficient}, dimentionality reduction methods, existing algorithms and techniques, evaluation methods and limitations \cite{kowsari2019text} and advantages based on applications. Paramesh \emph{et al} and Seongwook \emph{et al} compare the different classification algorithms such as multinomial naive bayes logistic regression, K-Nearest neighbour and Support Vector Machines (SVM) on real-world IT infrastructure ticket classifier system data, using different evaluation metrics in their research \cite{youn2007comparative,paramesh2019automated}. They claimed that SVM to have performed well on all the data samples. Random forest (RF) or naive bayes (NB) performed best in terms of correctly uncovering human intuitions. Hartmann \emph{et al} and his team present in their study that RF exhibits high performance in sentiment classification research done on  41 social media data sets covering major social media platforms, where the SVM never outperforms the RF \cite{hartmann2019comparing}. Cognitive RPA is efficiently undertaken as a low cost solution with Microsoft Azure Language Understanding Intelligent Service (LUIS) \cite{masood2019cognitive} and Azure machine learning studio.

Section III of this paper elaborates the process of automation. The section IV explains about the email classification approach, and the section V illustrates the results and their respective analysis. Finally, section VI contains the conclusion of the results.

\section{Method}
We are proposing a hybrid-process automation, in which we are introducing the automation architecture while adopting the manual process methodology. Incoming emails, that cannot be classified or understood by the knowledge base of the automation system will be sent for manual classification solution.

\subsection{Manual Process}
Providing technical support for large firms around the world has many challenges such as coordinating a vast amounts of mails and matching experts with employees who are in need of that expertise. When a technical issue is raised from a base level employee who works with applications, it is sent to the middle level and then to the higher level management of the respective regional branches throughout the hierarchical business architecture. Once it is approved by the branch manager, the issue email is forwarded to the technical coordinator to categorize the issue based on the priority level and technical requirements. Technical coordinator is responsible for the issues raised from the  regional branches all over the world.

Each regional branch is given a unique name such as  New York, Sydney, London, Beijing and Toronto mentioned as Category1 (cat1).  Category1 is identified by looking at the email address of the sender. Each regional branch has different plant applications that need different experts' consultation. Plant applications such as SAP, Darwin and infrastructure  are mentioned as Category2 (cat2). The possible plot of the issue emails such as computer, manufacturing, userID,  userunlock, financial, planning, purchasing issue generated by employees working in various plant applications across various regions are mentioned as Category3.

Mapping table is created with the plants placed in the regional offices and the issues created by the plants. Category1, Category2, Category3 contains  84, 8 and 77 unique categories to be classified. Table I shows some examples for each categories. Once all three categories are finalized by the technical coordinator, email tickets will be created and assigned to the admin-groups. Respective technical people in the admin-groups will provide consultancy and solve the issues. Not only one technician can handle issues assigned to many different admin groups allocated to him, but also particular admin category can be handled by many technicians as a group as well.

\begin{table}[htb!]
\caption{List of Categories}
\label{tab1}
\begin{tabular}{|c|c|c|c|} \hline
\textbf{Category1} & \textbf{\textit{Category2}}& \textbf{\textit{Category3}}& \textbf{\textit{AdminGroup}} \\ \hline
Newyork   & Darwin         & Planning   & Kandy-Darwin-Planning  \\
Battaya   & NonSAP         & Delivery   & Battay-NonSAP-Delivery \\
Colombo   & SAP            & UserUnlock & Colombo-Sap-UserUnlock \\
Colombo   & SAP            & UserID     & Colombo-sap-UserID     \\
City-1    & Infrastructure & Purchasing & Delhi-Infra-Purchasing \\ \hline
\end{tabular}
\end{table}

\subsection{Proposed Automation System}
\begin{figure}[htb!]
\includegraphics[height=8cm,width=8.7cm]{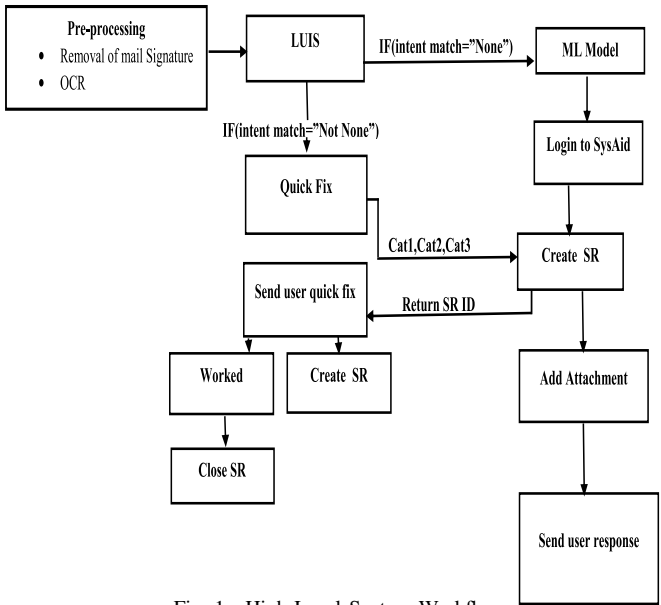}
\vspace{-0.5cm}
\caption{High-Level System  Workflow}
\label{fig1}
\end{figure}

In addition to replacing the technical coordinator role with AI bot to classify the raised email-issue tickets for respective admin groups, we propose instant quick fixes for some emails in an automated manner. High level workflow is described in Fig. 1. The AI bot has three main stages

\begin{itemize}
\item Quick fixes
\item Static rules
\item Email classifier
\end{itemize}

All the incoming mails are preprocessed for better quality of inputs. Signatures, greetings, Uniform Resource Locators (URL) are removed. Key body is extracted from the forwarded mails by digging deep into the mail contents. If an email contains attachments, Optical Character Recognition (OCR) is used to extract the text contents from the attachments.

\subsubsection{Quickfixes}
\begin{figure}[htb!]
\includegraphics[height=7cm,width=8.7cm]{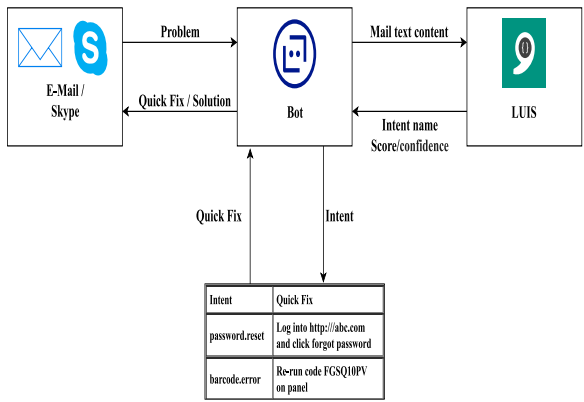}
\vspace{-1.0cm}
\caption{Quick fixes Bot architecture}
\label{fig2}
\end{figure}

Microsoft LUIS is used for instant quick fixes to provide solution based on prioritized emails. Fig. 2 shows the bot framework LUIS  architecture that handles the quick fixes. Quick fixes are trained with most occurring samples that need quick solutions. LUIS is a model that artificial intelligence applications use to predict the intention of phrases spoke. There are 3 main key phases categorized as defining phase, training phase and publishing phase. Natural language is extremely flexible with LUIS. Intents are the type of defined words that are supported by utterances. An action the user wants to perform can be defined by an intent. Fig. 3 elaborates the intent matching breakdown mechanism. Entities are identified form the sentences. Suitable entity will be selected for generating tickets.

\begin{figure}[htb!]
\centerline{\includegraphics[height=5cm,width=8cm]{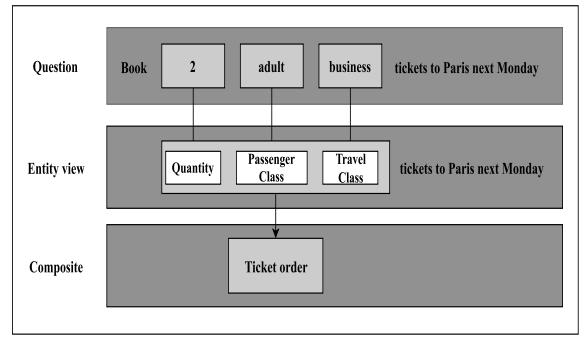}}
\vspace{-0.5cm}
\caption{LUIS intent matching}
\label{fig3}
\end{figure}

If an incoming email is identified with the matched intent, cat1, cat2, cat3 will be allocated. Tickets will be created for admin-groups. The issue will be solved using automated messages through a chat bot solution. If the issue is solved, then the ticket will be closed by the quick fixes. If it is too complicated for the knowledge of the BOT  then it creates ticket for adminGroup for the assistance of consultants.

\begin{figure}[htb!]
\includegraphics[height=9cm,width=8.7cm]{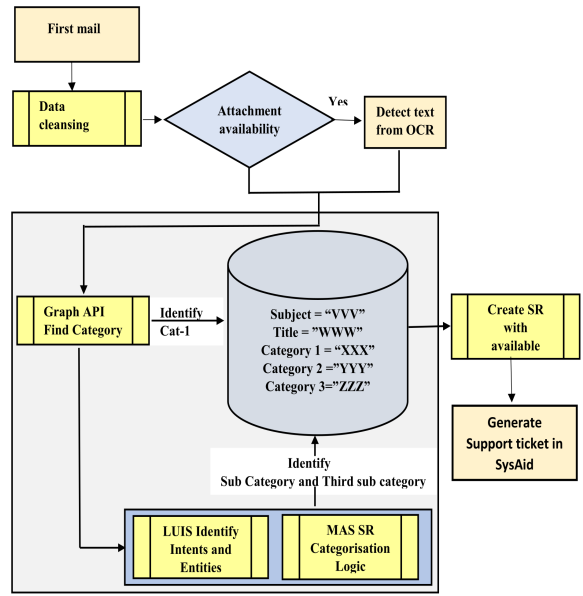}
\vspace{-1.0cm}
\caption{First mail flow}
\label{fig4}
\end{figure}

\begin{figure*}[htb!]\centering
\includegraphics[height=10cm,width=10cm]{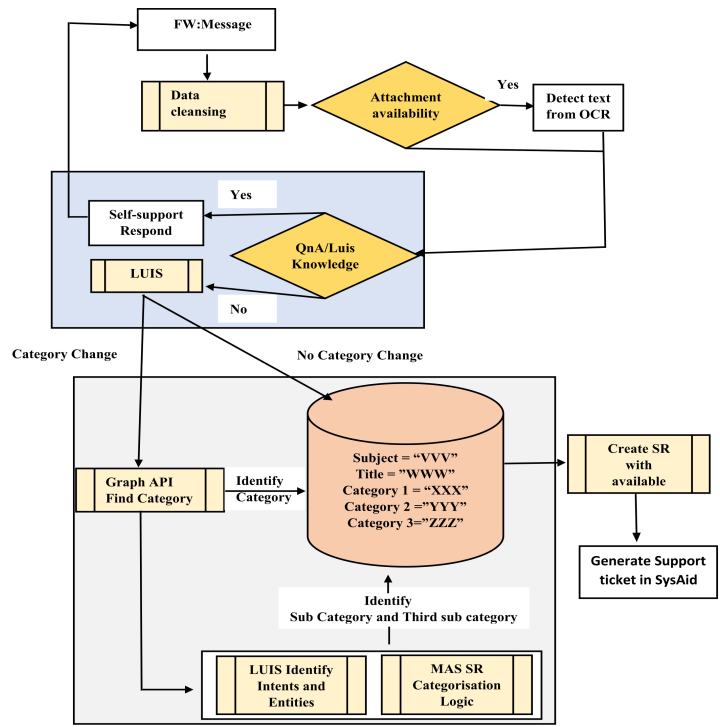}
\vspace{-0.5cm}
\caption{Forwarded mail flow with LUIS support}
\end{figure*}

The emails identified by static rules and keywords are classified with the highest accuracy. The knowledge base of static rules and keywords are gathered using feature engineering and the insights from the technical coordinator. Remaining emails are sent to a complex ensemble machine learning model to be classified.

Different types of emails are treated in a different way for efficient execution and to reduce the error.

\subsubsection{First mail}
\begin{figure*}[htb!]\centering
\includegraphics[height=12cm,width=14cm]{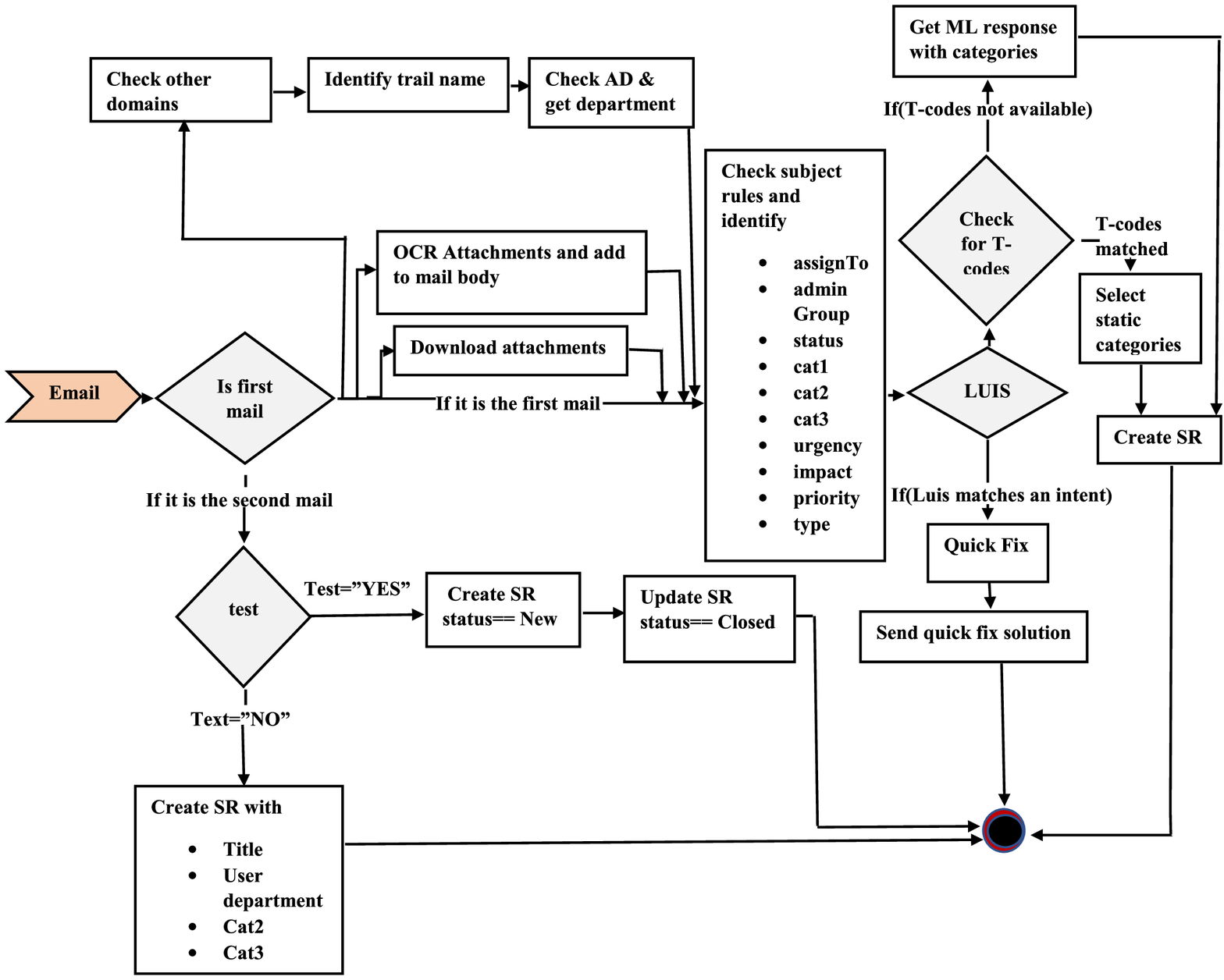}
\caption{Overall architecture}
\end{figure*}

Fig. 4 shows the flow of email categorization response for new incoming emails. If an incoming mail is a fresh new mail, it is initially subjected to cleaning. OCR will extract the texts from the attachment depending on the attachments' availability. Cat1 is assigned according to the knowledge of the database and sender details. According to the priority, emails are passed through LUIS. Thereafter if LUIS fails to solve the issue ML model will assign the cat2, cat3, Admin group for ticket creation.

\subsubsection{Forwarded mail}
If incoming mail is a continuation of previous email, it is directly handled by LUIS question and answer self automated support. Then it follows the normal procedure of categorization. Fig. 5 clearly illustrates the flow.

Fig. 6 explains the overall architecture. Static rules are mentioned as T-codes. Every categorized mails has to be assigned to respective consultant denoted as assignTo.

\section{ Email classifier using machine learning}
\subsection{Preprocessing}
Preprocessing is necessary to increase the accuracy of a text classification model, because it avoids the classification model focusing attention on unwanted sentences and intents. Emails are fed into Microsoft-Bot services. It handles the headers and outputs the processed channel-data in JavaScript Object Notation (JSON) format. The channel data summarizes the information such like sender, receiver, body, subject and important metadata. Regular expression (regex) can be used for searching strings by defining a search pattern. Regex findings are created to remove unwanted words from the channel data queries for further processing of the emails.

OCR has to be accurate in detecting text in an image. Microsoft-OCR is used for text recognition of this automation process. It extracts the recognized characters into a machine-usable character stream. Accuracy of the text recognition depends on the image quality such as blurry images, small text size, complex background, shadows and handwritten text. Since most of the image attachments are computer generated images and screen shots of error messages, Microsoft-OCR capabilities fits for the use case.

260000 emails are taken from past history.  Extracted preprocessed data from Microsoft-Bot and OCR services are saved as Comma-separated Values (CSV) files. It is further processed before feeding to machine learning model. Unwanted words are removed from the context using nltk library stopwords and manually collected stopwords. URLs, punctuation marks are removed. Every word is tokenized, lemmatized and normalized, i.e. title, body, OCR, from, to, CC, Cat1, Cat2, and Cat3.

\subsection{Feature selection}
Since the sender and receiver varies with time because of new employees' arrivals and old employees' resignations. In order to handle this fluctuating situation, To, CC, From columns are dropped from the input data. Cat1 is known from the email address. Cat2, Cat3 for specific cat1 is described in the table1. Cat2 and Cat3 are merged and defined as target category for classification. Nearly 180 custom features are created based on the plant's availability  and region mapping. It is embedded to understand the availability of plant and the issue for the given region denoted as Unique-Category. Based on mapping table (extension of table1), custom features ensures that whether the plant application (cat2) and the technical issue (cat3) belongs to the regional plant (cat1).

By the analysis made from the existing samples and from the human semantic knowledge of the technical coordinator, it is sensed that not only the title of the email is enough to predict the category, but also the attachment and body play a major role.

\subsection{Machine learning approach}
Even though labelled data set was provided, initially unsupervised learning algorithm K-Nearest Neighbor (KNN) clustering was applied to the data set to observe the possibility of clusters \cite{yong2009improved}. Since number of unique categories of the target field (Unique-Cat) is 77, there are many common words between categories. It is too confusing and not showing promising categories and accuracies. Multi class multi label classification supervised algorithms such as random forest, XGBoost are used as benchmarks.

\subsubsection{Feature selection}
\begin{itemize}
\item Ngrams
are a continuous sequence of n items from a given sample of text. From title, body and OCR text words are selected. Ngrams of 3 nearby words are extracted with Term Frequency-Inverse Document Frequency (TF-IDF) vectorizing, then features are filtered using chi squared the feature scoring method.
\item Feature hashing
is a method to extract the features from the text. It allows to use variable size of feature vectors with standard learning algorithms. 12000 features are hashed from the text, OCR and title. Then using chi-squared statistical analysis 200 best features that fits with target unique-category are selected.
\end{itemize}

\subsubsection{Random forest}
Random Forest is a bagging Algorithm, an ensemble learning  method for classification that operates by constructing a multitude of decision trees at training time and outputting the class that has highest mean majority vote of the classes\cite{pal2005random}.

\subsubsection{XGBoost}
XGBoost is a decision-tree-based ensemble Machine Learning algorithm that uses a gradient boosting framework. It is used commonly in the classification problems involving unstructured data\cite{stein2019analysis}.

\subsubsection{Hierarchical Model}
Since the number of target labels are high, achieving the higher accuracy is difficult, while keeping all the categories under same feature selection method. Some categories performs well with lower TF-IDF vectorizing range and higher n grams features even though they showed lower accuracy in the overall single model. Therefore, hierarchical machine learning models are built to classify 31 categories in the first classification model and remaining categories are named as low-accu and predicted as one category. In the next model, predicted low-accu categories are again classified into 47 categories. Comparatively this hierarchical model works well since various feature selection methods are used for various categories\cite{stein2019analysis}.

\subsection{Deep learning approach}
\subsubsection{LSTM}
Long short term memory is an artificial neural network architecture which outperforms most of the  machine learning algorithms. In the deep learning approach, feature selection is done in neurons weight matrix by itself. Bidirectional long short term memory (LSTM) is used with glove word embedding to predict the categories\cite{lee2016sequential}.

\subsubsection{BERT}
Even though Bert is the state of the art model, for the considered data set it hasn't shown-up with the maximum breach of accuracy for the expected automation\cite{devlin2018bert}. When we consider the commercial model for the inference, having a dedicated Kubernetes cluster with high performance computer is costly. So complex models with high computation power are not considered as abetter solution.

\subsection{Threshold  Selection}
In order to classify only higher confident emails, the thresholds for each and every 73 categories are defined. For an incoming email, the probability of assigning each category will be calculated. Best category will be selected based on the maximum probability out of those 73 probabilities. By looking at overall F-score, thresholding decisions are made. For the low accuracy categories (accuracy  less than 75 percentage) higher threshold level is set. For middle accuracy categories (accuracy less than 90 percentage) min probability of correctly classified samples are taken. Higher accuracy categories (accuracy  greater than 90 percentage) are left free with 0 threshold to classify all the incoming emails. The threshold techniques as a bottle neck decreases the number of samples classified by the autonomous process, but it increases the accuracy of the classified samples. The proposed thresholds satisfy the expected manual workload reduction as well as the accuracy percentage.

In this paper Randomforest, XGBoost, LSTM, Bidirectional LSTM  with embeddings are analyzed with different input features. Complex deep-learning models such like transformers are not used in order to go for low cost inference solution. Train set and test set are divided as 80:20 percentage. Precision, recall, F-score are taken as  evaluation metrics.

\section{Results  and Analysis}
Automation of quick email replies for technical queries increase the overall efficiency of day to day processes by 3 percentage. Even though replacing the manual Human email-assigner entirely with AI bot is not possible, yet the automation ML model handles 61 percentage of incoming emails correctly. It is reducing massive human effort per day. For generalization purpose email's title, body, attachments are considered in increasing accuracy, while ignoring sender, receiver, carbon copy information. Table II shows the accuracy percentages for different models with different feature selection methods. An accuracy of 77.3 percentage was obtained without any thresholding techniques for 73 classes multiclasss multi label classification problem. With threshold adjustments for each categories, it was increased to 85.6 percentage. Increasing threshold values results in reducing the number of mails classified by ML-model. It is necessary to handle limited number of high confident emails by the ML-model due to ensure the promising accuracy levels. Feature Engineering for custom feature selection and, Hierarchical cascade modelling increases the accuracy of the XGBoost machine learning model to reach accuracy of the LSTM models. By cascading model1 (mod1) with 83.2 accuracy for 31 classes and model2 (mod2) with 71.1 accuracy for 47 low-accuracy classes, overall hierarchical model exhibited 76.5 accuracy. All the accuracy terms refers F-score. Selected keywords were used as static rules accurate classification. Since accuracy is considerably satisfactory for the automation process, the system was deployed. The incorrectly classified mails are handled manually after the proper notification by the technical consultant.

\begin{table}[htb!]\centering
\caption{Model accuracy}
\begin{tabular}{|c|c|c|c|} \hline
\textbf{Input Features} & \textbf{\textit{selection}}& \textbf{\textit{Model}}& \textbf{\textit{Accuracy}} \\ \hline
Title               & Feature & RF-ensemble      & 48.6      \\
                    & Hashing & XG-Boost         & 48.6      \\
                    &         & LSTM             & 52.1      \\
                    &         & LSTM + embedding & 52.3      \\ \hline
Title + Body        & Feature & RF-ensemble      & 62.9      \\
                    & Hashing & XG-Boost         & 63.8      \\
                    &         & LSTM             & 67.1      \\
                    &         & LSTM + embedding & 68.4      \\ \hline
Title + Body + OCR  & Feature & RF-ensemble      & 67.9      \\
                    & Hashing & XG-Boost         & 69.1      \\
                    &         & LSTM             & 73.2      \\
                    &         & LSTM+embedding   & 74.6      \\ \hline
Body+Title+OCR      & Feature & Hierarchical ML  & 76.5      \\
with                & hashing &                  & Mod1 83.2 \\
Custom Data         &         &                  & Mod2 71.1 \\
Engineered Features &         & Lstm+embedding   & 77.3      \\ \hline
\end{tabular}
\end{table}
\begin{figure*}[htbp]\centering
\includegraphics[height=6.5cm,width=16cm]{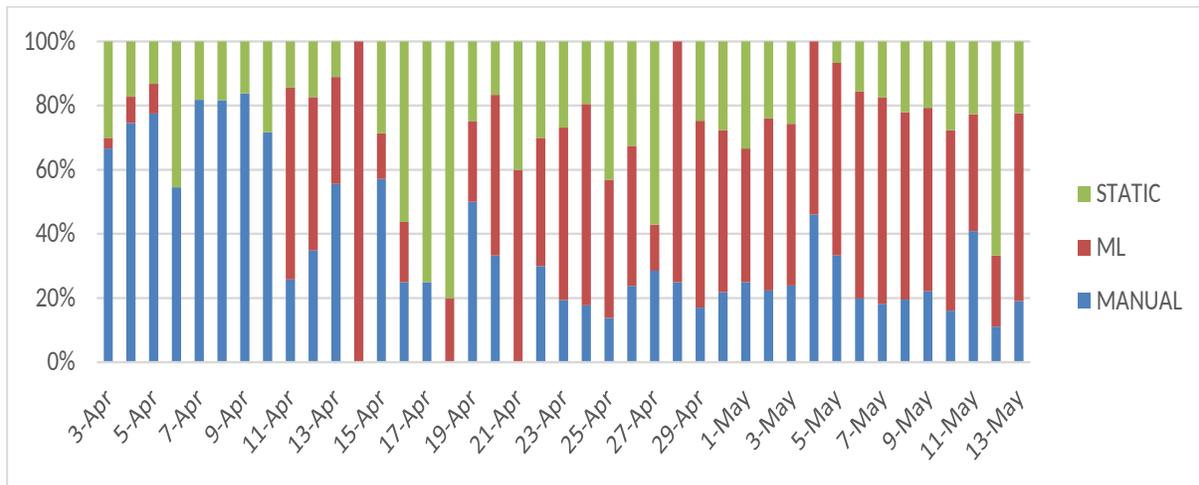}
\caption{Daily solved emails}
\label{fig7}
\end{figure*}
Fig. 7 Shows emails classified by the ML, static rules and manual process represented in daily basis. Incoming emails per day varies between 30 to 120. It clearly illustrates the effect of retraining. After 10-April, the percentages of emails classified per day was increased as well as accuracy.

Fig. 8 shows average monthly analysis of incoming mails after each retraining. Average Monthly incoming mails are  calculated as 1467 per month by considering a 4 months period. Initial training  was done on august 2018 with 170,000 samples, model was able to classify nearly 50 percentage of incoming emails. After the second retraining on january 2019 with 200,000 sample, model classified 58 percentage of incoming mails per month. Third retraining was done on April 2019 with 260000 samples. Results stated that nearly 61 percentage of incoming mails were handled by ML model. Nearly 20 percentage of incoming emails were handled by static rules. Automation bot was proved to handle 81 percentage of the total incoming mails per month including ML and static rules, leading to efficient human-machine interaction, Instant problem solving and fast process.

\begin{figure}[htb!]\centering
\includegraphics[height=4cm,width=8.7cm]{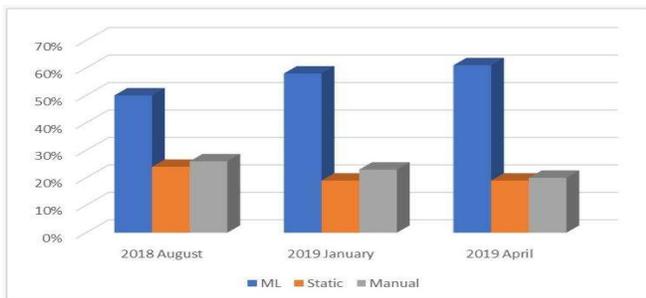}
\caption{Monthly efficiency.}
\label{fig8}
\end{figure}

\section{Conclusion}
Quick fixes from Microsoft LUIS Bot framework provides instant solutions for the raised email queries. Input text features of emails such as title, body, attachment OCR text and the feature engineered custom features all together outperform for the considered real word email data set. Sure-shot Static rules and hierarchical machine learning model with statistically calculated threshold enhances the accuracy of the overall system to an acceptable percentage. Bidirectional LSTM with word embedding techniques are implemented finally with thresholding techniques. Less complex Machine learning models lead to low cost virtual machine solutions for serving. Robotic Process Automation Architecture reduces human effort of email support desk by 81 percentage while having a reasonable accuracy of 85.6 percentage.

\bibliographystyle{IEEEtran}
\bibliography{ref}
\end{document}